\documentclass[unnumsec,webpdf,contemporary,large]{oup-authoring-template}%
\graphicspath{{Fig/}}
\theoremstyle{thmstyleone}%
\theoremstyle{thmstyletwo}%
\theoremstyle{thmstylethree}%
\usepackage{subfiles}
\usepackage{bibunits}
\usepackage{amsmath, amssymb, amsthm, bm, amsfonts}
\usepackage{booktabs}

\definecolor{mycustompurple}{RGB}{154, 36, 79} % 定义自己的颜色
\usepackage[utf8]{inputenc}
\usepackage{framed}
\hypersetup{
    colorlinks=true,        % 激活链接颜色，去掉链接边框
    linkcolor=blue,         % 文档内部链接颜色（如图表等引用）
    citecolor=blue,         % 文献引用链接颜色
    filecolor=blue,         % 文件链接颜色
    urlcolor=blue           % 外部URL链接颜色
}

\begin{document}
\journaltitle{Briefings in Bioinformatics}
\DOI{DOI HERE}
\copyrightyear{2024}
\pubyear{2014}
\access{Advance Access Publication Date: Day Month Year}
\appnotes{Paper}

\firstpage{1}
%\subtitle{Subject Section}

\title[SpaDiT]{SpaDiT: Diffusion Transformer for Spatial Gene Expression Prediction using scRNA-seq}

\author[1$^\#$]{Xiaoyu Li}
\author[2$^\#$]{Fangfang Zhu}
\author[1$^\ast$]{Wenwen Min\ORCID{0000-0002-2558-2911}}

\authormark{Xiaoyu Li et al.}
\address[1]{\orgdiv{the School of Information Science and Engineering}, \orgname{Yunnan University}, \orgaddress{ \postcode{650091}, \state{Kunming, Yunnan}, \country{China}}}

\address[2]{\orgdiv{College of Nursing Health Sciences}, \orgname{Yunnan Open University},  \orgaddress{\postcode{650599}, \state{Kunming}, \country{China}}}

\corresp[$\ast$]{Corresponding author:\href{email:email-id.com}{minwenwen@ynu.edu.cn} and $^\#$Co-first authors}

\received{Date}{0}{Year}
\revised{Date}{0}{Year}
\accepted{Date}{0}{Year}

% \editor{Associate Editor: Name}
% \abstract{
% \textbf{Motivation:} .\\
% \textbf{Results:} .\\
% \textbf{Availability:} .\\
% \textbf{Contact:} \href{name@email.com}{name@email.com}\\
% \textbf{Supplementary information:} Supplementary data are available at \textit{Journal Name}
% online.}

\abstract{
The rapid development of spatial transcriptomics (ST) technologies is revolutionizing our understanding of the spatial organization of biological tissues. Current ST methods, categorized into next-generation sequencing-based (seq-based) and fluorescence in situ hybridization-based (image-based) methods, offer innovative insights into the functional dynamics of biological tissues. However, these methods are limited by their cellular resolution and the quantity of genes they can detect.
To address these limitations, we propose SpaDiT, a deep learning method that utilizes a diffusion generative model to integrate scRNA-seq and ST data for the prediction of undetected genes. By employing a Transformer-based diffusion model, SpaDiT not only accurately predicts unknown genes but also effectively generates the spatial structure of ST genes.
We have demonstrated the effectiveness of SpaDiT through extensive experiments on both seq-based and image-based ST data. SpaDiT significantly contributes to ST gene prediction methods with its innovative approach. Compared to eight leading baseline methods, SpaDiT achieved state-of-the-art performance across multiple metrics, highlighting its substantial bioinformatics contribution.
% Our code is available at \url{https://github.com/lllxxyyy-lxy/SpaDiT}.
% Source code and all datasets used in this paper are available at \url{https://github.com/wenwenmin/XXXXX} and \url{https://zenodo.org/records/XXXXX}.
}
\keywords{diffusion model, spatial transcriptomics data, scRNA-seq data, Transformer}

\maketitle
\section{Introduction}
Single-cell RNA sequencing (scRNA-seq) can represent the entire transcriptome of a specific cell in an organ, providing an excellent perspective for in-depth study of various behaviors and mechanisms between cells \cite{intro-sc1}. 
However, since scRNA-seq must undergo sample tissue dissociation, it also leads to the inability of scRNA-seq to capture the spatial distribution and spatial information of cells, which is often crucial for understanding the complex physiological processes between cells \cite{intro-sc2}. 
Therefore, spatial transcriptomics (ST) has emerged as an advanced technology that can retain spatial location information while measuring gene expression in tissue or cell samples \cite{intro-st1}. 
This technology enables researchers to parse the spatial distribution of gene expression in tissues, enhancing the understanding of cell types, functions, interactions, and key details in development, disease, and biological processes.

At present, ST technology can be mainly divided into two categories: Based on next-generation sequencing technology (seq-based): such as 10x Visium \cite{10Xvisium}, Slide-seq \cite{slide-seq} and Stereo-seq \cite{stereo-seq}, transcriptome-wide gene expression within a spatial point can be detected. 
Fluorescence in situ hybridization (image-based): such as seqFish \cite{seqFish} and MERFISH \cite{Merfish}, can measure thousands of genes at the resolution of single cells, but they usually lack full transcriptome coverage, resulting in only a few hundred genes in actual sequencing. Although these two technologies can detect gene expression in the whole transcriptome range, their capture rate is low due to their resolution \cite{st-weak1,st-weak2}. 
The current solution mainly focuses on increasing the capture rate and predicting uncaptured genes by using scRNA-seq data to enhance ST data to improve its resolution \cite{st-method1,st-method2,st-method3}.

In recent years, a variety of methods have been proposed to use scRNA-seq data to improve the resolution of ST data and predict uncaptured genes. These methods, such as Tangram \cite{Tangram}, scVI \cite{scVI}, SpaGE \cite{SpaGE}, stplus \cite{stPlus}, SpaOTsc \cite{SpaOTsc}, novoSpaRc \cite{novoSpaRc}, SpatialScope \cite{SpatialScope}, stDiff \cite{stDiff}. 
They all assume that scRNA-seq data and ST data have similar expression distributions, and they identify the similarity between scRNA-seq cells and ST cells by detecting the expression patterns of shared genes. 
Then, these methods use similar scRNA-seq cells to predict the unmeasured part of ST data. However, due to the sparse nature of scRNA-seq and ST data, and the reliance on common genes to calculate similarity, this poses a huge challenge to how to align the two data. 
In addition, simply using scRNA-seq as a reference for ST data prediction is difficult to avoid introducing batch bias of scRNA, which increases the difficulty of predicting unknown genes \cite{st-method-weak}. 

In this paper, we introduce a novel method named SpaDiT, which uses a conditional diffusion model to understand and generate unmeasured gene expression in ST data.
Although diffusion models have made significant contributions in the field of computer vision and have shown excellent performance in the field of protein or drug generation \cite{ddpm2015,cdm-cv1,dm-drug}, their application in genomics is still relatively limited. 
The goal of SpaDiT is to utilize scRNA-seq as a prior input in the diffusion model to help the model understand the relationship between gene expressions, thereby guiding the model to generate genes that are not measured in ST data. SpaDiT utilizes genes in single cells as unique identifiers by incorporating them in the diffusion model along with the corresponding genes in ST, and employs the Transformer-based diffusion model to enhance the model's prediction accuracy of specific genes.

We conduct a comprehensive performance evaluation on 10 ST datasets based on different sequencing technologies, different tissues, and different sample sizes, and compare them with the current state of the art (SOTA) methods. 
The results show that our model achieves the best performance on all five evaluation indicators, and the correlation between predicted gene expression and actual gene expression shows the best accuracy. 
This shows that SpaDiT can effectively make predictions when predicting unmeasured gene expression in ST data. 
In addition, the genes predicted by our model have a high spatial similarity with the genes in the actual ST data. For the spatial expression patterns of each data set, our model can accurately predict and clearly divide the spatial boundaries. 
This demonstrates SpaDiT's ability to predict ST data gene expression and provide subsequent analysis.

\section{Materials and methods}
\subsection{\textbf{Datasets and pre-processing}}

%%%%%%%%%%%%%%%%%%%%%%%%%%%%%%table1
\begin{table*}[]
\centering
\caption{The list of ten paired scRNA-seq and spatial transcriptomic datasets. The first five ST datasets are image-based, the next five datasets are sequencing-based. HPR: hypothalamic preoptic region; PMC: primary motor cortex.}
\label{data_description}
\resizebox{\textwidth}{!}{%
\begin{tabular}{l|l|l|ll|cc|cc|cc|cc|cc}
\hline
 &
  \multicolumn{1}{c|}{} &
   &
  \multicolumn{2}{c|}{\textbf{Platform}} &
  \multicolumn{2}{c|}{\textbf{Number of Cells/Spots}} &
  \multicolumn{2}{c|}{\textbf{Number of Genes}} &
  \multicolumn{2}{c|}{\textbf{Prepro. Cells/Spots}} &
  \multicolumn{2}{c|}{\textbf{Prepro. Genes}} &
  \multicolumn{2}{c}{\textbf{Dropout Rate}} \\ \cline{4-15} 
\multirow{-2}{*}{\textbf{Datasets}} &
  \multicolumn{1}{c|}{\multirow{-2}{*}{\textbf{Tissue}}} &
  \multirow{-2}{*}{\textbf{GEO ID}} &
  \multicolumn{1}{c|}{SC} &
  \multicolumn{1}{c|}{ST} &
  \multicolumn{1}{c|}{SC (Cells)} &
  ST (Spots) &
  \multicolumn{1}{c|}{SC (Genes)} &
  \multicolumn{1}{c|}{ST (Genes)} &
  \multicolumn{1}{c|}{SC (Cells)} &
  ST (Spots) &
  \multicolumn{1}{c|}{SC (Genes)} &
  \multicolumn{1}{c|}{ST (Genes)} &
  \multicolumn{1}{c|}{SC} &
  ST \\ \hline
MH \cite{data-MH} &
  mouse hippocampus &
  GSE158450 &
  \multicolumn{1}{l|}{10X Chromium} &
  seqFish &
  \multicolumn{1}{c|}{8596} &
  3585 &
  \multicolumn{1}{c|}{16384} &
  249 &
  \multicolumn{1}{c|}{8584} &
  {\color[HTML]{000000} 3585} &
  \multicolumn{1}{c|}{1260} &
  249 &
  \multicolumn{1}{c|}{80.3\%} &
  6.3\% \\
MHPR \cite{data-MHPR} &
  mouse HPR &
  GSE113576 &
  \multicolumn{1}{l|}{10X Chromium} &
  MERFISH &
  \multicolumn{1}{c|}{31299} &
  4975 &
  \multicolumn{1}{c|}{18646} &
  154 &
  \multicolumn{1}{c|}{31297} &
  4975 &
  \multicolumn{1}{c|}{1939} &
  153 &
  \multicolumn{1}{c|}{73.7\%} &
  62.2\% \\
ML \cite{data-MG} &
  mouse liver &
  GSE109774 &
  \multicolumn{1}{l|}{Smart-seq2} &
  seqFISH &
  \multicolumn{1}{c|}{981} &
  2177 &
  \multicolumn{1}{c|}{17533} &
  19532 &
  \multicolumn{1}{c|}{887} &
  2177 &
  \multicolumn{1}{c|}{2279} &
  569 &
  \multicolumn{1}{c|}{73.2\%} &
  75.4\% \\
MG \cite{data-MG} &
  mouse gastrulation &
  GSE15677 &
  \multicolumn{1}{l|}{10X Chromium} &
  seqFISH &
  \multicolumn{1}{c|}{4651} &
  8425 &
  \multicolumn{1}{c|}{19103} &
  351 &
  \multicolumn{1}{c|}{4651} &
  8425 &
  \multicolumn{1}{c|}{1945} &
  345 &
  \multicolumn{1}{c|}{58.6\%} &
  74.1\% \\
MVC \cite{data-MVC} &
  mouse visual cortex &
  - &
  \multicolumn{1}{l|}{Smart-seq} &
  STARmap &
  \multicolumn{1}{c|}{14249} &
  1549 &
  \multicolumn{1}{c|}{34041} &
  1020 &
  \multicolumn{1}{c|}{14249} &
  1549 &
  \multicolumn{1}{c|}{3774} &
  844 &
  \multicolumn{1}{c|}{58.2\%} &
  76.2\% \\ \cline{1-15} 
MHM \cite{data-MHM} &
  mouse hindlimb muscle &
  GSE161318 &
  \multicolumn{1}{l|}{10X Chromium} &
  10X Visium &
  \multicolumn{1}{c|}{4816} &
  995 &
  \multicolumn{1}{c|}{15460} &
  33217 &
  \multicolumn{1}{c|}{4809} &
  995 &
  \multicolumn{1}{c|}{1667} &
  416 &
  \multicolumn{1}{c|}{80.3\%} &
  68.9\% \\
HBC \cite{data-HBC} &
  human breast cancer &
  CID3586 &
  \multicolumn{1}{l|}{10X Chromium} &
  10X Visium &
  \multicolumn{1}{c|}{6178} &
  4784 &
  \multicolumn{1}{c|}{21164} &
  28402 &
  \multicolumn{1}{c|}{6143} &
  4784 &
  \multicolumn{1}{c|}{625} &
  125 &
  \multicolumn{1}{c|}{76.6\%} &
  70.6\% \\
ME \cite{data-ME} &
  mouse embryo &
  GSE160137 &
  \multicolumn{1}{l|}{10X Chromium} &
  10X Visium &
  \multicolumn{1}{c|}{3415} &
  198 &
  \multicolumn{1}{c|}{19374} &
  53574 &
  \multicolumn{1}{c|}{3415} &
  198 &
  \multicolumn{1}{c|}{2163} &
  540 &
  \multicolumn{1}{c|}{61.1\%} &
  62.3\% \\
MPMC \cite{data-MPMC} &
  mouse PMC &
  - &
  \multicolumn{1}{l|}{10X Chromium} &
  10X Visium &
  \multicolumn{1}{c|}{3499} &
  9852 &
  \multicolumn{1}{c|}{24340} &
  24518 &
  \multicolumn{1}{c|}{3499} &
  9852 &
  \multicolumn{1}{c|}{2544} &
  636 &
  \multicolumn{1}{c|}{70.6\%} &
  81.7\% \\
MC \cite{data-MC}&
  mouse cerebellum &
  SCP948 &
  \multicolumn{1}{l|}{10X Chromium} &
  Slide-seqV2 &
  \multicolumn{1}{c|}{26252} &
  41674 &
  \multicolumn{1}{c|}{24409} &
  23264 &
  \multicolumn{1}{c|}{26252} &
  41674 &
  \multicolumn{1}{c|}{822} &
  205 &
  \multicolumn{1}{c|}{79.5\%} &
  83.9\% \\ \hline
\end{tabular}%
}
\end{table*}

In this paper, we collected ten benchmark datasets (scRNA sequencing and spatial transcriptomics data) from different tissues of various organisms. 
As illustrated in \autoref{data_description}, these datasets originate from various biological organizations and utilize differing sequencing platforms and technologies. They also vary in sample sizes, number of spatially measured genes, and missing data rates. 
Specifically, the sequencing platforms for single-cell data in these datasets include 10X Chromium, Smart-seq, and Smart-seq2. For spatial transcriptomics data, the platforms are seqFISH, MERFISH, 10X Visium, STARmap, and Slide-seqV2. 
These datasets are derived from different biological tissues, primarily from mouse and human breast cancer tissue sections.

For the implementation of SpaDiT, we adhered to the data preprocessing protocols as established in prior studies \cite{Benchmark}. 
More specifically, we first removed genes with no expression from both the single-cell and spatial transcriptomics datasets. 
Subsequently, we screened the remaining genes to identify those that were highly expressed, using criteria based on the number of genes in each dataset.

We partitioned the processed data into training, validation, and test sets with ratios of 7:2:1, respectively. 
These subsets are mutually independent, with the test set being strictly separate from the training set. 
All reported results were derived solely from evaluations on the test set.

\subsection{\textbf{The architeture of SpaDiT}}
\begin{figure*}
    \centering
    \includegraphics{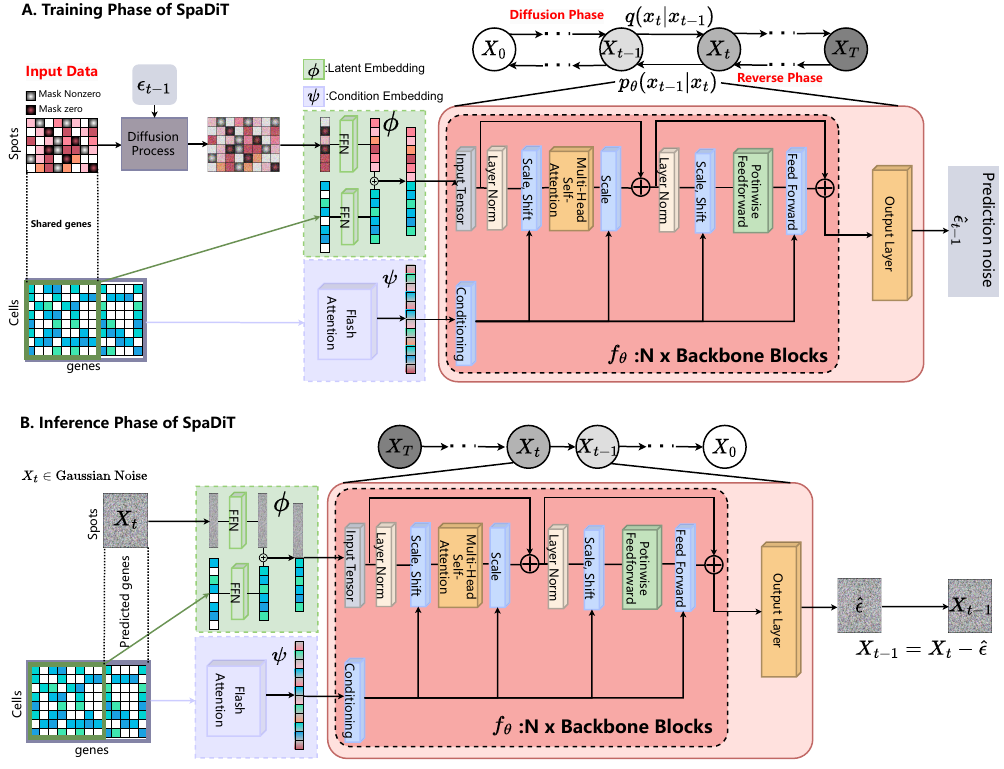}
    \caption{The architecture of SpaDiT. There are three parts in total: latent embedding, conditional embedding and network backbone. 
    (A) is the training process where each gene is considered as a sample, and (B) is the inference process.
    }
    \label{model}
\end{figure*}
SpaDiT is a conditional diffusion-based deep generative model that enhances spatial transcriptomics data by leveraging single-cell RNA sequencing (scRNA-seq) data as prior information, aiming to accurately predict the expression of unmeasured or unknown genes.
As illustrated in the \autoref{model}, SpaDiT takes two types of input data: a gene expression matrix from spatial transcriptomics data and another from scRNA-seq data. Utilizing a conditional diffusion model, SpaDiT uses scRNA-seq data as a conditioning factor to guide the model through the diffusion and denoising processes, thereby generating the targeted gene expression profiles for the spatial transcriptomic data. The SpaDiT architecture comprises three key modules: the Latent Embedding module for processing spatial transcriptomic data, the Condition Embedding module for processing scRNA-seq data, and the core network architecture: Diffusion with Transformer, which facilitates the integration and generation of data.
In the following sections, we will introduce the main modules of SpaDiT.

\subsubsection{\textbf{Latent Embedding in SpaDiT}}
In the proposed SpaDiT, the latent embedding module is crucial. Instead of operating directly on real data, we work within an efficient, low-dimensional latent space, which is better suited for likelihood-based generative models. 
Therefore, we utilize an encoder to map the high-dimensional input data to a low-dimensional representation, and we train the diffusion model within this latent space. 

Notably, our proposed method involves two types of data input: spatial transcriptomics data ($X_{\text{st}}$) and scRNA-seq data ($X_{\text{sc}}$). The genes in $X_{\text{sc}}$ are divided into shared genes ($G_{\text{share}}$) with spatial transcriptomics data and unique genes ($G_{\text{unique}}$). 
For the input of Latent Embedding, we define it as follows: for each sample (i,e., gene) \( x_{\text{st}}^i \in X_{\text{st}} \) in the spatial transcription data, we integrate it with the scRNA-seq data \( x_{\text{sc}}^i \in X_{\text{sc}} \), where \( x_{\text{st}}^i \) and \( x_{\text{sc}}^i \) in \( G_{\text{share}} \) are utilized as inputs. In latent embedding, we employ a simple feed-forward network to project \( x_{\text{st}}^i \) and \( x_{\text{sc}}^i \) into the same dimensional space, concatenating the projected \( \hat{x}_{\text{st}}^i \) and \( \hat{x}_{\text{sc}}^i \) as the output of the latent embedding, that is, \( x_\phi = \hat{x}_{\text{st}}^i \oplus \hat{x}_{\text{sc}}^i \).

\subsubsection{\textbf{Condition Embedding in SpaDiT}}
The condition embedding module leverages scRNA-seq data as a conditioning factor in our model, integrating it into the diffusion process to guide the model in generating the required gene expression. Given that scRNA-seq data is high-dimensional, directly using the entire matrix as input would result in the curse of dimensionality. Consequently, the condition embedding module utilizes an attention mechanism to convert the high-dimensional single-cell data matrix into a lower-dimensional representation. This reduces the data to a low-dimensional, high-expression latent representation, which is then used as a conditional mechanism in subsequent diffusion model training.

For the input of Condition Embedding, the high-dimensional input matrix \( X_{\text{sc}} \) is processed using Flash-attention to compute a lower-dimensional representation \( X_{\psi} \) as output:
\begin{equation}
\begin{aligned}
    Q &= X_{\text{sc}} W^Q, K = X_{\text{sc}} W^K, V = X_{\text{sc}} W^V, \\
    X_{\psi} &= \text{softmax}\left(\frac{Q \Phi_K(K)^T}{\sqrt{d_k}}\right) \Phi_V(V)
\end{aligned}
\end{equation}
Where:
\begin{itemize}
    \item \( W^Q \), \( W^K \), and \( W^V \) are projection matrices that transform \( X \) into queries \( Q \), keys \( K \), and values \( V \), respectively.
    \item \( \Phi_K \) and \( \Phi_V \) are the dimensionality reduction functions applied to \( K \) and \( V \), resulting in lower-dimensional.
    \item \( d_k \) is the dimension of \( K \) after projection, used to scale the softmax computation.
    \item The \(\text{softmax}\) function is applied over each row, normalizing the dot product scores into a probability distribution used to compute the weighted sum of values \(  \Phi_V(V) \).
\end{itemize}

\subsubsection{\textbf{Diffusion with Transformer in SpaDiT}}
The backbone network of our proposed SpaDiT is Diffusion Transformers (DiTs), a new architecture for diffusion models. 
For the backbone network model, we refer to previous work \cite{DiT} and make modifications based on the challenges we encounter. 
Our backbone model has two types of input: $x_\phi$, representing latent embedding, and $x_\psi$, representing condition embedding. We initialize each residual block in the backbone network as an identity function and incorporate the condition embedding into the backbone. 
At each layer, we also perform scaling regression on all residual connections within the backbone, facilitating rapid model convergence.

After the final DiT block, the gene expression token sequence needs to be decoded into output noise prediction and output diagonal covariance prediction. 
The shapes of both outputs are identical to the input in the original space, and a standard linear decoder is employed to achieve this. 
Finally, the decoded tokens are rearranged to match the layout of the original expression, yielding the predicted noise and covariance.

\subsubsection{\textbf{Training phase in SpaDiT}}
Here in, SpaDiT works with two types of input data: the spatial transcriptomics data $X_{\text{st}} = \{ x_{\text{st}}^{i} \}_i^n \in \mathbb{R} ^ {n \times p}$ and scRNA-seq data $X{\text{sc}} = \{ x_{\text{sc}}^{j} \}_j^m \in \mathbb{R} ^ {m \times q}$. Among them, $n$ and $p$ respectively represent the number of genes and the number of spots in spatial transcriptomics data, and $m$ and $q$ respectively represent the number of genes and the number of cells in scRNA-seq data.

The training phase of SpaDiT is shown in the \autoref{model}~(A). 
We first mask the genes of the original spatial transcriptomics data according to a certain proportion, where the mask is divided into two parts: the part with an expression value of zero and the part with an expression value that is not zero. 
The input tensor of the training phase is defined as follows:
\begin{equation}
    x_0 = \hat{x}_{\text{st}}^i \oplus \hat{x}_{\text{sc}}^i
    \label{def_x0}
\end{equation}
where  \( \hat{x}_{\text{st}}^i \) and \( \hat{x}_{\text{sc}}^i \) are projection of  \( x_{\text{st}}^i \) and \( x_{\text{sc}}^i \) by a simple feed-forward network.

In the realm of DDPMs \cite{ddpm2015,ddpm2020}, consider the task of learning a model distribution \( p_\theta(\mathbf{x}_0) \) that closely approximates a given data distribution \( q(\mathbf{x}_0) \). 
Suppose we have a sequence of latent variables \( \mathbf{x}_t \) for \( t = 1, \ldots, T \), existing within the same sample space as \( \mathbf{x}_0 \), which is denoted as \( \mathcal{X} \). 
DDPMs are latent variable models that are composed of two primary processes: the forward process and the reverse process. 
The forward process is defined by a Markov chain, as follows:
\begin{equation}
  q(\mathbf{x}_{1:T} | \mathbf{x}_0) := \prod_{t=1}^{T} q(\mathbf{x}_t | \mathbf{x}_{t-1}),
\label{eq_ddpm_forward}
\end{equation}
where $q(\mathbf{x}_t | \mathbf{x}_{t-1}) := \mathcal{N}(\sqrt{1 - \beta_t}\mathbf{x}_{t-1}$, 
and the variable \(\beta_t\) is a small positive constant indicative of a noise level. The sampling of \(x_t\) can be described by the closed-form expression \(q(x_t | x_0) = \mathcal{N}(x_t; \sqrt{\alpha_t}x_0, (1 - \alpha_t)\mathbf{I})\), where \(\hat{\alpha}_t := 1 - \beta_t\) and \(\alpha_t\) is the cumulative product \(\alpha_t := \prod_{i=1}^t \hat{\alpha}_i\). Consequently, \(x_t\) is given by the equation \(x_t = \sqrt{\alpha_t}x_0 + (1 - \alpha_t)\epsilon\), with \(\epsilon \sim \mathcal{N}(0, \mathbf{I})\). In contrast, the reverse process aims to denoise \(x_t\) to retrieve \(x_0\), a process which is characterized by the ensuing Markov chain:
\begin{equation}
\begin{aligned}
&p_\theta(\mathbf{x}_{0:T}) := p(\mathbf{x}_T) \prod_{t=1}^{T} p_\theta(\mathbf{x}_{t-1} | \mathbf{x}_t), \quad \mathbf{x}_T \sim \mathcal{N}(0, \mathbf{I}), \\
&p_\theta(\mathbf{x}_{t-1} | \mathbf{x}_t) := \mathcal{N}(\mathbf{x}_{t-1}; \mu_\theta(\mathbf{x}_t, t), \sigma^2_\theta(\mathbf{x}_t, t)\mathbf{I}),  \\
& \mu_\theta(\mathbf{x}_t, t) = \frac{1}{\alpha_t} \left( \mathbf{x}_t - \frac{\beta_t}{\sqrt{1-\alpha_t}} \epsilon_\theta(\mathbf{x}_t, t) \right), \\
& \sigma_\theta(\mathbf{x}_t, t) = \beta_t^{1/2} %, \\
\end{aligned}
\label{eq_ddpm_reverse}
\end{equation}
where $\epsilon_\theta(\mathbf{x}_t, t)$ is a trainable denoising function and
\begin{equation}
\beta_t = 
\begin{cases} 
\frac{1-\bar{\alpha}_{t-1}}{1-\bar{\alpha}_t} \beta_1, & \text{for } t > 1, \\
\beta_1, & \text{for } t = 1.
\end{cases} 
\end{equation}

SpaDiT aims to help models understand and estimate the expression of missing genes in ST data by utilizing scRNA-seq data as prior information, thereby enabling the model to better predict gene expression from ST data.
We represent the data of the condition as $x_{0}^c = x_{\psi}$.
Therefore, our goal is to estimate the posterior $p((E_{n,1} - m_1) \odot ((E_{n,1} - m_2) \odot x_{0}))| x_{0}^c)$, where $E_{n, 1}$ is an all-1 matrix $n\times 1$ with dimension , $m_1, m_2 \in \{ 0, 1 \}^{n \times 1} $ is an element-wise indicator, representing the zero and non-zero parts of the mask respectively.

We also denote predicted genes as $x_{t}^{*}$, where t is the time step.
Therefore, the goal of our SpaDiT conditional mechanism is to estimate the probability:
\begin{equation}
    p_\theta(x_{t-1}^{*} | x_{t}^{*}, x_{0}^c).
\end{equation}
In order to better use the scRNA-seq data as a priori conditions for the diffusion model to perform prediciting gene expression, 
% we transform the Eq. (\ref{eq_ddpm_forward}) and Eq. (\ref{eq_ddpm_reverse}) into:
we transform the \autoref{eq_ddpm_forward} and \autoref{eq_ddpm_reverse} into:
\begin{equation}
\small
\begin{split}
&p_\theta(x_{0:T}^{*} | x_{ 0}^c) := p(x_{T}^{*}) \prod_{t=1}^{T} p_\theta(x_{t-1}^{*} | x_{ t}^{*}, x_{ 0}^c), x_{ T}^{*} \sim \mathcal{N}(0, \mathbf{I}). \\
&p_\theta(x_{t-1}^{*} | x_{t}^{*}, x_{ 0}^c) := \mathcal{N}(x_{ t-1}^{*}; \mu_\theta(x_{ t}^{*}, t | x_{0}^c), \sigma_\theta(x_{t}^{*}, t | x_{0}^c)\mathbf{I}).
\end{split}
\label{eq_spadit_forward_reverse}
\end{equation}

We can optimize the \autoref{eq_spadit_forward_reverse} parameters by minimizing the variational lower bound:
\begin{equation}
\small
\mathbb{E}_{q}\left[ -\log p_{\theta}(x_{0} \mid x_{ 0}^c) \right] \leq \mathbb{E}_{q}\left[ -\log \frac{p_{\theta}(x_{ 0:T} \mid x_{0}^c)}{q(x_{ 1:T} \mid x_{0})} \right].
\label{elbo_conditon}
\end{equation}

Also we can get a simplified training objective:
\begin{equation}
     \mathbb{E}_{x_{0}\sim q(\mathbf{x}_{0}), \epsilon\sim\mathcal{N}(0,\mathbf{I}),t}\|(\epsilon - \epsilon_{\theta}(\mathbf{x}_{t}^*, t | x_{ 0}^c))\|^2_2 .
    \label{eq11}
\end{equation}

We provide the training procedure of SpaDiT in Algorithm~\ref{algo_training}.
\begin{algorithm}
\caption{Training of SpaDiT}
\label{algo_training}
\begin{algorithmic}[1]
\State \textbf{Input:} ST data $X_{\text{st}} = \{ x_{st}^{i} \}_i^n \in \mathbb R ^ {n \times p}$, SC data $X_{\text{sc}} = \{ x_{sc}^{j} \}_j^m \in \mathbb R ^ {m \times q}$, Number of iterations $N_{\text{iter}}$, $\{\alpha_t\}_{t=1}^T$, $T$
\State \textbf{Output:} Trained denoising function $\epsilon_\theta$
\For{$i = 1$ to $N_{\text{iter}}$}
    \State $x_i \sim X_{\text{st}},~ x_j \sim X_{\text{sc}}$
    \State $x_0 = \Phi(x_i, x_j), ~x_0^c = \psi (X_{\text{sc}}), ~\text{where} [i,j] \in (X_\text{st} \cap X_\text{sc})$
    \State $t \sim \text{Uniform}(\{1, \dots, T\})$
    \State $\epsilon \sim \mathcal{N}(0, I)$
    \State Take gradient step on 
    \[
    \nabla_{\theta} \|\left(\epsilon - \epsilon_\theta(\sqrt{\alpha_t}{x}_{0}^* + \sqrt{1-\alpha_t}\epsilon, t| x_0^c)\right) \|^2_2
    \]
\EndFor
\end{algorithmic}
\end{algorithm}

\subsubsection{\textbf{Inference phase in SpaDiT}}
We focus on improving the conditional diffusion model characterized by the inverse process described in \autoref{eq_spadit_forward_reverse}. 
Our goal is to accurately model the conditional distribution $p\left(x_{t-1}^{*} | x_{t}^{*}, x_{ 0}^{c}\right )$ without resorting to approximations . 
To achieve this, we adapt the parameterization of DDPM from \autoref{eq_ddpm_reverse} for the conditional setting.
We introduce a conditional denoising function $\epsilon_{\theta} : \left(\mathcal{X}^* \times \mathbb{R} \mid \mathcal{X}^c\right) \rightarrow \mathcal{X }^ *$ accepts conditional observation value $x_{ 0}^c$ as input parameter.
On this basis, we use $\epsilon_{\theta}$ for parameterization, as follows:
\begin{equation}
\begin{split}
\mu_{\theta}(x_{t}^*,t | x_{ 0}^c) &= \mu \left(x_{ t}^{*}, t, s_{\theta}\left(x_{ t}^{*}, t | x_{ 0}^c\right)\right),\\
\sigma_{\theta}(x_{ t}^*, t | x_{ 0}^c) &= \sigma \left(x_{ t}^*, t\right),
\end{split}
\label{eq10}
\end{equation}
where $\mu$ and $\sigma$ are defined in \autoref{eq_ddpm_reverse}.
Utilizing the function $\epsilon_{\theta}$ and the data $x_0$, we can simulate samples of $x_{ 0}^*$ by employing the reverse process outlined in \autoref{eq_spadit_forward_reverse}. 
We provide the inference procedure of SpaDiT in Algorithm~\ref{algo_inference}.
\begin{algorithm}
\caption{Inference of SpaDiT}
\label{algo_inference}
\begin{algorithmic}[1]
\State \textbf{Input:} Gaussian Noise $\mathcal{N}(0, I)$, SC data $X_{\text{sc}} = \{ x_{sc}^{i} \}_i^m \in \mathbb R ^ {m \times q}$
\State \textbf{Output:} Predicted gene expression $x_{0}$
\For{$t = T$ to $1$}
    \State $x_t^c = \psi (X_{\text{sc}})$
    \State Sample $x_t, \epsilon_t \sim \mathcal{N}(0, I)$ 
    \State $x_t= \Phi(x_{t}, x_{\text{sc}}^i ), ~\text{where} [i] \in (X_\text{st} \cap X_\text{sc})$
    \State $x_{t-1} \leftarrow \frac{1}{\sqrt{\alpha_t}} \left( x_{t} - \frac{1-\alpha_t}{\sqrt{1-\alpha_t}} \epsilon_\theta(x_{t}, t | x_t^c) \right) + \sqrt{\beta_t} \varepsilon_t$
    \State $t \leftarrow t - 1$
\EndFor
\end{algorithmic}
\end{algorithm}

\subsection{\textbf{Evaluation metrics}} \label{metrics}
To evaluate the performance of SpaDiT and other baseline methods, we use five evaluation indicators: Pearson Correlation Coefficient (PCC), Structural Similarity Index Measure (SSIM), Root Mean Square Error (RMSE), Jensen-Shannon Divergence (JS) and Accuracy Score (AS) to evaluate the gene prediction performance of different methods on ten datasets. The specific definition of the evaluation metrics can be found in \href{https://github.com/lllxxyyy-lxy/SpaDiT}{Supplementary Materials}.

\subsection{\textbf{Baselines}}
\begin{figure*}
    \centering
    \includegraphics[width=\textwidth]{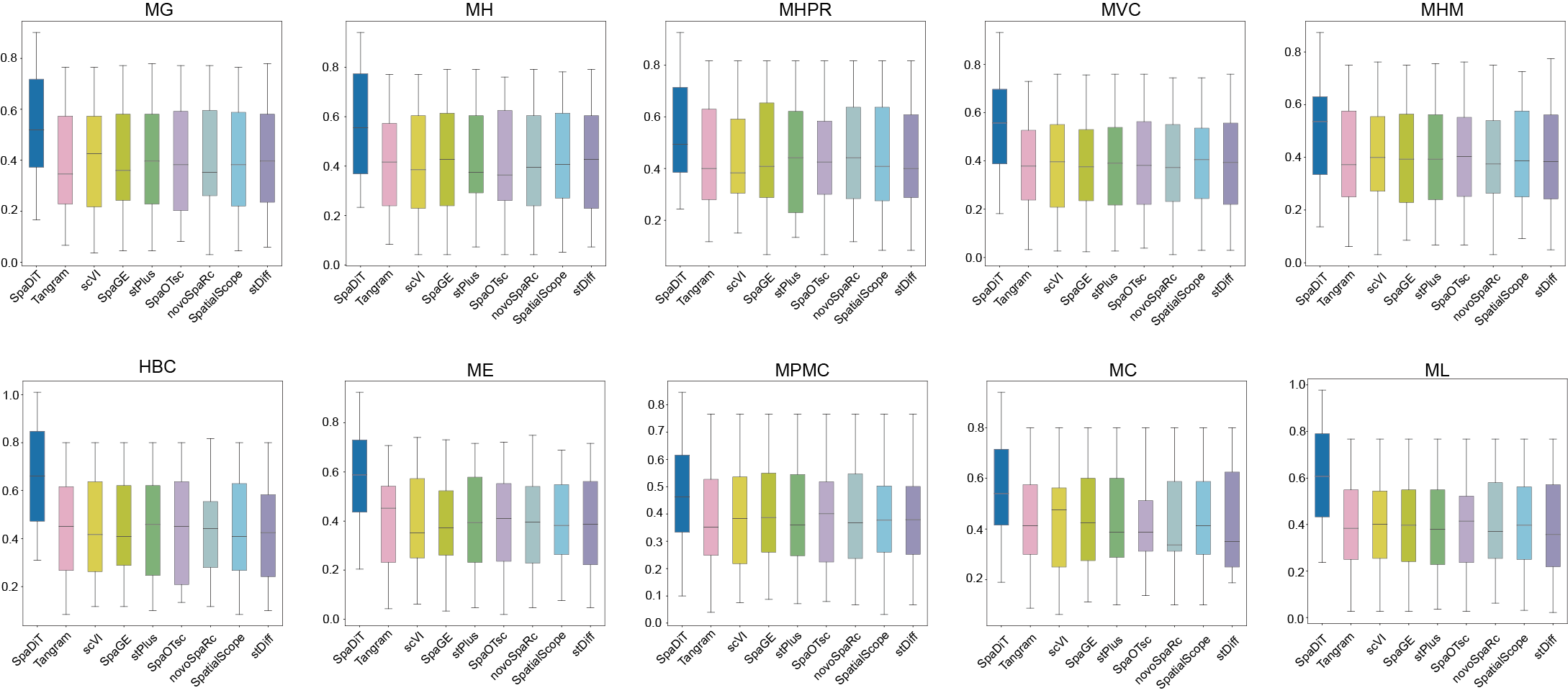}
    \caption{Performance evaluation is based on the comprehensive metric of Accuracy Score (AS) on ten real paired ST and scRNA-seq datasets. Accuracy Score (AS) is a comprehensive indicator for evaluating model performance. The definition can be found in \autoref{metrics}. The central line represents the median, the box depicts the interquartile range, whiskers extend to 1.5 times the interquartile range, and dots represent the AS of individual datasets.}
    \label{AS}
\end{figure*}
We compared the performance of SpaDiT to eight baseline methods, with data processing procedures (e.g., normalization and scaling) consistent for each method.
The specific baselines are as follows:
\begin{itemize}
    \item Tangram \cite{Tangram}: It is a method that can map any type of sc/snRNA-seq data, including multimodal data such as those from SHARE-seq, which can be used to reveal spatial patterns of chromatin accessibility. We refer to the guide on the Tangram GitHub repository: \url{https://github.com/broadinstitute/Tangram}.

    \item scVI \cite{scVI}: It is a scalable framework for probabilistic representation and analysis of single-cell gene expression. We refer to the guide on the scVI GitHub repository: \url{https://github.com/YosefLab/scVI}.

    \item SpaGE \cite{SpaGE}: It is a method that integrates spatial and scRNA-seq datasets to predict whole-transcriptome expressions in their spatial configuration. We refer to the guide on the SpaGE GitHub repository: \url{https://github.com/tabdelaal/SpaGE}.

    \item stPlus \cite{stPlus}: It is a reference-based method that leverages information in scRNA-seq data to enhance spatial transcriptomics. We refer to the guide on the stPlus GitHub repsitory: \url{https://github.com/xy-chen16/stPlus}.

    \item SpaOTsc \cite{SpaOTsc}: It is a method that relies on structured optimal transfer to recover the spatial properties of scRNA-seq data by exploiting spatial measurements of a relatively small number of genes. We refer to the guide on the SpaOTsc GitHub repository: \url{https://github.com/zcang/SpaOTsc}.

    \item novoSpaRc \cite{novoSpaRc}: It is a method that reconstructs tissue based on this hypothesis and optionally improves the reconstruction by including a reference map of marker genes. We refer to the guide on the SpaOTsc GitHub repository: \url{https://github.com/rajewsky-lab/novosparc}.

    \item SpatialScope \cite{SpatialScope}: It is a method to integrate scRNA-seq reference data and ST data using deep generative models. We refer to the guide on the SpatialScope GitHub repository: \url{https://github.com/YangLabHKUST/SpatialScope}.

    \item stDiff \cite{stDiff}: It is a method that capturing gene expression abundance relationships in scRNA-seq data through two Markov processes. We refer to the guide on the stDiff GitHub repository: \url{https://github.com/fdu-wangfeilab/stDiff}
\end{itemize}

\section{Results}
\subsection{\textbf{SpaDiT improves prediction accuracy of spatial gene expression}}
To rigorously assess the SpaDiT method's capabilities in predicting gene expression, we conducted a comparative analysis with eight other widely recognized methods in the field. We used four key performance metrics, as outlined in \autoref{metrics}, to systematically evaluate both SpaDiT and the comparator baseline methods. The evaluation focused on computing the mean and variance of these metrics across all genes within each dataset. The results are depicted in \autoref{baseline}.

Our findings indicate that SpaDiT consistently achieves state-of-the-art (SOTA) performance in all four metrics across the ten examined datasets. However, it is important to note that in a few cases, SpaDiT slightly lags behind some established methods in one particular metric. This deviation provides critical insights into scenarios where SpaDiT might be further optimized.

In addition to these traditional metrics, we introduced an advanced scoring system, referred to as AS metrics, to further evaluate SpaDiT’s performance. The results, illustrated in \autoref{AS}, confirm that SpaDiT not only meets but often exceeds the performance benchmarks set by the baseline methods across all ten spatial transcriptomics (ST) datasets. The inclusion of AS metrics provides a more nuanced understanding of SpaDiT's predictive prowess, underscoring its robustness and effectiveness in diverse experimental conditions. This comprehensive approach solidifies SpaDiT’s position as a leading method in gene expression prediction, highlighting its potential to significantly enhance the accuracy and reliability of spatial transcriptomics analyses.

%%%%%%%%%%%%%%%%%%%%%%%%%%%%%%%%%%%%%%%%%%%%%%%%Baseline
\begin{table*}[]
\centering
\caption{Comparison with baseline methods on the ten paired scRNA-seq and ST datasets.}
\label{baseline}
\resizebox{\textwidth}{!}{%
\begin{tabular}{l|c|c|c|c|c|c|c|c|c|c}
\hline
PCC$\uparrow$ &
  MG &
  MH &
  MHPR &
  MVC &
  MHM &
  HBC &
  ME &
  MPMC &
  MC &
  ML \\ \hline
Tangram \cite{Tangram} &
  0.458±0.203 &
  0.523±0.116 &
  0.683±0.012 &
  0.623±0.117 &
  0.536±0.053 &
  0.703±0.142 &
  0.503±0.025 &
  0.727±0.026 &
  0.745±0.003 &
  0.714±0.056 \\
scVI \cite{scVI} &
  0.476±0.157 &
  0.446±0.157 &
  0.691±0.143 &
  0.594±0.023 &
  0.511±0.117 &
  0.656±0.005 &
  0.496±0.007 &
  0.716±0.014 &
  0.736±0.015 &
  0.637±0.001 \\
SpaGE \cite{SpaGE}&
  0.526±0.114 &
  0.438±0.163 &
  0.653±0.063 &
  0.603±0.107 &
  0.545±0.226 &
  0.639±0.025 &
  0.512±0.013 &
  0.753±0.066 &
  0.769±0.011 &
  0.653±0.007 \\
stPlus \cite{stPlus}&
  0.503±0.233 &
  0.401±0.037 &
  0.483±0.231 &
  0.574±0.059 &
  0.476±0.007 &
  0.597±0.111 &
  0.526±0.026 &
  0.689±0.007 &
  0.701±0.099 &
  0.699±0.014 \\
SpaOTsc \cite{SpaOTsc}&
  0.522±0.014 &
  0.485±0.107 &
  0.657±0.002 &
  0.629±0.147 &
  0.496±0.018 &
  0.587±0.107 &
  0.547±0.006 &
  0.734±0.201 &
  0.738±0.064 &
  0.723±0.005 \\
novoSpaRc \cite{novoSpaRc}&
  0.563±0.158 &
  0.567±0.252 &
  0.613±0.146 &
  0.656±0.037 &
  0.515±0.003 &
  0.647±0.122 &
  0.569±0.013 &
  0.756±0.015 &
  0.756±0.015 &
  0.766±0.056 \\
SpatialScope \cite{SpatialScope}&
  0.612±0.143 &
  0.582±0.183 &
  0.637±0.031 &
  0.683±0.114 &
  0.547±0.103 &
  0.733±0.183 &
  0.563±0.056 &
  0.769±0.022 &
  0.776±0.006 &
\textbf{0.803±0.014} \\
stDiff \cite{stDiff}&
  0.482±0.021 &
  0.527±0.013 &
  0.621±0.007 &
  0.601±0.043 &
  0.471±0.009 &
  0.544±0.021 &
  0.553±0.014 &
  0.629±0.011 &
  0.604±0.019 &
  0.736±0.099 \\
\textbf{SpaDiT (Ours)} &
  \textbf{0.657±0.035} &
  \textbf{0.621±0.099} &
  \textbf{0.770 ±0.043} &
  \textbf{0.725±0.106} &
  \textbf{0.573±0.083} &
  \textbf{0.772±0.057} &
  \textbf{0.590±0.146} &
  \textbf{0.808±0.043} &
  \textbf{0.812±0.039} &
  0.784±0.096 \\ \hline \noalign{\medskip} \hline
SSIM$\uparrow$ &
  MG &
  MH &
  MHPR &
  MVC &
  MHM &
  HBC &
  ME &
  MPMC &
  MC &
  ML \\ \hline
Tangram \cite{Tangram}&
  0.355±0.114 &
  0.541±0.203 &
  0.681±0.025 &
  0.653±0.115 &
  0.388±0.109 &
  0.656±0.007 &
  0.521±0.047 &
 \textbf{0.889±0.043} &
  0.789±0.004 &
  0.689±0.005 \\
scVI \cite{scVI}&
  0.487±0.155 &
  0.422±0.128 &
  0.647±0.121 &
  0.564±0.025 &
  0.374±0.115 &
  0.617±0.028 &
  0.587±0.013 &
  0.674±0.012 &
  0.736±0.006 &
  0.694±0.014 \\
SpaGE \cite{SpaGE}&
  0.503±0.003 &
  0.403±0.158 &
  0.631±0.011 &
  0.611±0.004 &
  0.401±0.006 &
  0.588±0.189 &
  0.513±0.064 &
  0.653±0.011 &
  0.667±0.055 &
  0.703±0.023 \\
stPlus \cite{stPlus}&
  0.533±0.114 &
  0.367±0.127 &
  0.657±0.176 &
  0.656±0.007 &
  0.426±0.013 &
  0.638±0.221 &
  0.479±0.023 &
  0.627±0.103 &
  0.693±0.011 &
  0.736±0.014 \\
SpaOTsc \cite{SpaOTsc}&
  0.547±0.126 &
  0.503±0.013 &
  0.701±0.026 &
  0.637±0.021 &
  0.484±0.170 &
  0.626±0.118 &
  0.601±0.188 &
  0.663±0.114 &
  0.718±0.004 &
  0.688±0.007 \\
novoSpaRc \cite{novoSpaRc}&
  0.587±0.028 &
  0.537±0.026 &
  0.713±0.123 &
  0.631±0.018 &
  0.477±0.201 &
  0.633±0.107 &
  0.622±0.023 &
  0.726±0.055 &
  0.726±0.006 &
  0.705±0.006 \\
SpatialScope \cite{SpatialScope}&
  0.612±0.016 &
  0.588±0.014 &
  0.731±0.054 &
  0.674±0.026 &
  \textbf{0.512±0.122} &
  0.659±0.055 &
  \textbf{0.701±0.022} &
  0.826±0.014 &
  0.753±0.014 &
  0.714±0.003 \\
stDiff \cite{stDiff}&
  0.463±0.017 &
  0.548±0.118 &
  0.673±0.013 &
  0.576±0.007 &
  0.462±0.017 &
  0.514±0.012 &
  0.563±0.017 &
  0.598±0.019 &
  0.701±0.023 &
  0.688±0.017 \\
\textbf{SpaDiT (Ours)} &
  \textbf{0.632±0.037} &
  \textbf{0.574±0.125} &
  \textbf{0.738±0.044} &
  \textbf{0.689±0.114} &
  0.495±0.175 &
  \textbf{0.717±0.111} &
  0.688±0.144 &
  0.781±0.050 &
  \textbf{0.787±0.042} &
  \textbf{0.751±0.107} \\ \hline \noalign{\medskip} \hline
RMSE$\downarrow$ &
  MG &
  MH &
  MHPR &
  MVC &
  MHM &
  HBC &
  ME &
  MPMC &
  MC &
  ML \\ \hline
Tangram \cite{Tangram}&
  1.263±0.053 &
  1.412±0.018 &
  1.263±0.012 &
  1.587±0.041 &
  1.237±0.005 &
  1.542±0.003 &
  1.633±0.004 &
  1.324±0.048 &
  1.216±0.184 &
  1.346±0.015 \\
scVI \cite{scVI}&
  1.155±0.012 &
  1.363±0.026 &
  1.374±0.026 &
  1.327±0.106 &
  1.213±0.103 &
  1.378±0.005 &
  1.581±0.013 &
  1.207±0.034 &
  1.179±0.067 &
  1.411±0.056 \\
SpaGE \cite{SpaGE}&
  1.187±0.025 &
  1.433±0.037 &
  1.287±0.029 &
  1.354±0.047 &
  1.347±0.025 &
  1.413±0.101 &
  1.553±0.024 &
  1.137±0.011 &
  1.213±0.005 &
  1.233±0.008 \\
stPlus \cite{stPlus}&
  1.254±0.003 &
  1.367±0.045 &
  1.384±0.121 &
  1.289±0.022 &
  1.156±0.014 &
  1.331±0.077 &
  1.496±0.033 &
  1.656±0.007 &
  1.154±0.024 &
  1.303±0.014 \\
SpaOTsc \cite{SpaOTsc}&
  1.433±0.058 &
  1.213±0.058 &
  1.203±0.027 &
  1.253±0.007 &
  1.227±0.058 &
  1.203±0.114 &
  1.403±0.004 &
  1.227±0.026 &
  1.016±0.007 &
  1.263±0.005 \\
novoSpaRc \cite{novoSpaRc}&
  1.275±0.143 &
  1.526±0.213 &
  1.252±0.011 &
  1.206±0.014 &
  1.412±0.117 &
  1.198±0.007 &
  1.556±0.021 &
  1.334±0.015 &
  0.967±0.153 &
  1.523±0.007 \\
SpatialScope \cite{SpatialScope}&
  1.019±0.022 &
  1.288±0.258 &
  1.201±0.003 &
 \textbf{1.009±0.007} &
  1.217±0.005 &
  1.102±0.005 &
  1.483±0.007 &
  1.104±0.056 &
 \textbf{0.863±0.004}&
  1.343±0.014 \\
stDiff \cite{stDiff}&
  1.326±0.019 &
  1.325±0.022 &
  1.081±0.013 &
  1.219±0.066 &
  1.312±0.007 &
  1.217±0.023 &
  1.561±0.023 &
  1.326±0.016 &
  1.224±0.003 &
  1.223±0.009 \\
\textbf{SpaDiT (Ours)} &
  \textbf{0.877±0.049} &
  \textbf{1.103±0.015} &
  \textbf{1.184±0.058} &
  1.116±0.038 &
  \textbf{1.125±0.060} &
  \textbf{0.992±0.045} &
  \textbf{1.376±0.118} &
  \textbf{1.089±0.038} &
  1.004±0.037 &
  \textbf{1.121±0.047} \\ \hline \noalign{\medskip} \hline
JS$\downarrow$ &
  MG &
  MH &
  MHPR &
  MVC &
  MHM &
  HBC &
  ME &
  MPMC &
  MC &
  ML \\ \hline
Tangram \cite{Tangram}&
  0.477±0.057 &
  0.254±0.003 &
  0.458±0.033 &
\textbf{0.343±0.007} &
  0.502±0.056 &
  0.397±0.105 &
  0.803±0.026 &
  0.403±0.056 &
  0.547±0.005 &
  0.347±0.014 \\
scVI \cite{scVI}&
  0.426±0.088 &
  0.324±0.147 &
  0.496±0.011 &
  0.403±0.001 &
  0.537±0.113 &
  0.427±0.089 &
  0.749±0.015 &
  0.423±0.115 &
  0.601±0.014 &
  0.363±0.047 \\
SpaGE \cite{SpaGE}&
  0.437±0.054 &
  0.272±0.023 &
  0.511±0.007 &
  0.387±0.114 &
  0.528±0.007 &
  0.415±0.026 &
  0.882±0.003 &
  0.374±0.004 &
  0.617±0.006 &
  0.403±0.011 \\
stPlus \cite{stPlus}&
  0.481±0.146 &
  0.288±0.057 &
  0.503±0.014 &
  0.399±0.005 &
  0.488±0.125 &
  0.439±0.005 &
  0.814±0.036 &
  0.393±0.005 &
  0.576±0.004 &
  0.423±0.016 \\
SpaOTsc \cite{SpaOTsc}&
  0.513±0.126 &
  0.334±0.058 &
  0.411±0.022 &
  0.403±0.147 &
  0.503±0.111 &
  0.411±0.015 &
  0.792±0.007 &
  0.417±0.011 &
  0.463±0.026 &
   \textbf{0.311±0.007} \\
novoSpaRc \cite{novoSpaRc}&
  0.488±0.003 &
  0.401±0.017 &
  0.389±0.005 &
  0.412±0.003 &
  0.496±0.015 &
  0.429±0.085 &
  0.683±0.015 &
  0.401±0.005 &
  0.431±0.005 &
  0.401±0.006 \\
SpatialScope \cite{SpatialScope}&
  0.403±0.002 &
  0.263±0.174 &
  0.366±0.007 &
  0.389±0.008 &
  0.487±0.026 &
  0.455±0.002 &
  0.622±0.150 &
  0.389±0.107 &
  0.407±0.014 &
  0.355±0.014 \\
stDiff \cite{stDiff}&
  0.467±0.001 &
  0.412±0.015 &
  0.387±0.021 &
  0.461±0.011 &
  0.467±0.021 &
  0.456±0.011 &
  0.663±0.017 &
  0.436±0.022 &
  0.432±0.063 &
  0.396±0.007 \\
\textbf{SpaDiT  (Ours)} &
  \textbf{0.346±0.012} &
  \textbf{0.246±0.005} &
  \textbf{0.337±0.010} &
  0.369±0.029 &
  \textbf{0.463±0.116} &
  \textbf{0.381±0.061} &
  \textbf{0.549±0.134} &
  \textbf{0.356±0.012} &
  \textbf{0.371±0.013} &
  0.421±0.064 \\ \hline
\end{tabular}%
}
\end{table*}

\subsection{\textbf{SpaDiT enhances the similarity of predicted gene expression in high-dimensional space}}
%UMAP
\begin{figure*}
    \centering
    \includegraphics[width=1\textwidth]{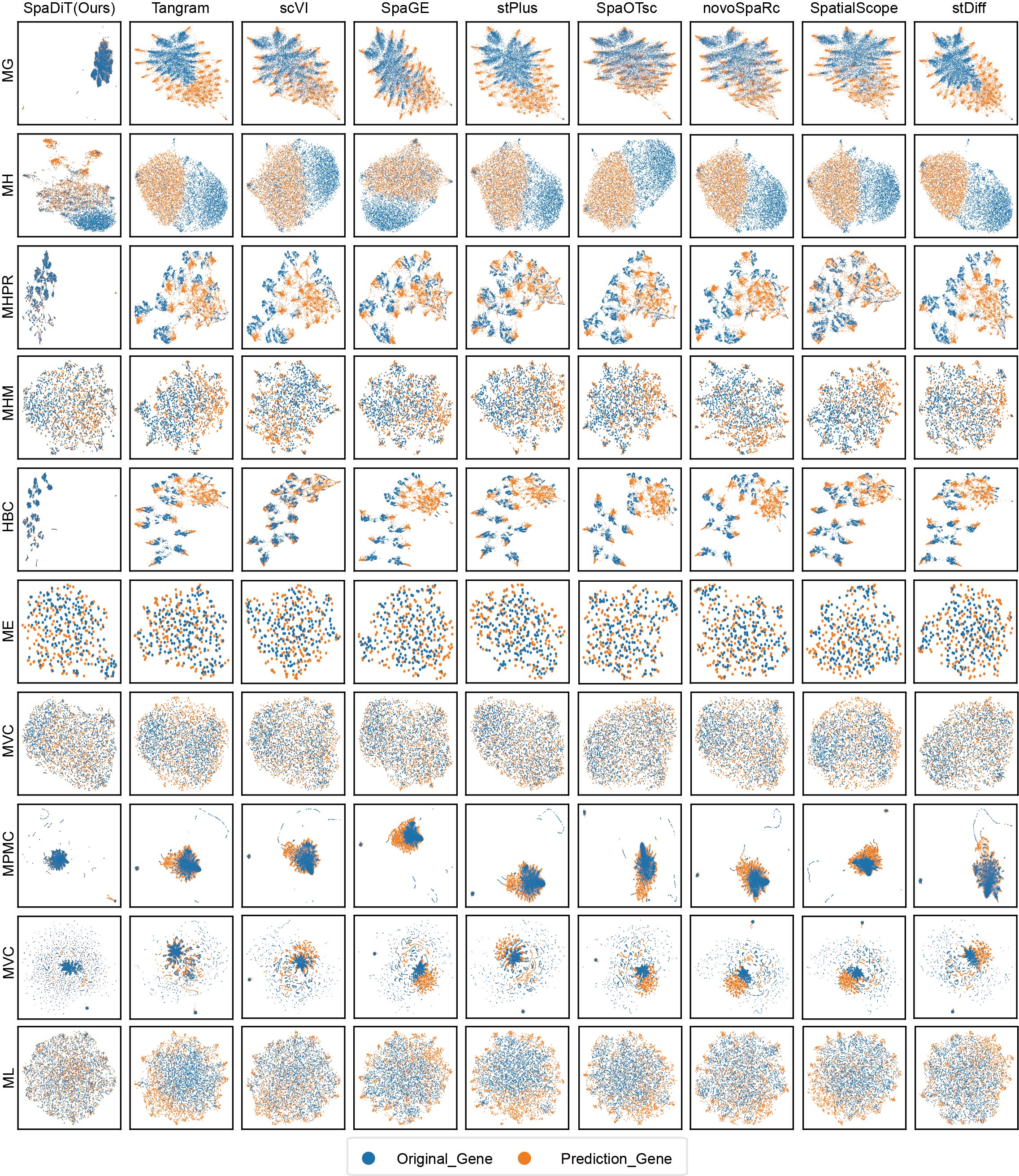}
    \caption{UMAP plots illustrating gene predicted by SpaDiT,Tangram, scVI,  SpaGE, stPlus, SpaOTsc, novoSpaRc, SpatialScope and stDiff. The closer the two scatter points are, the better the prediction effect is. The scatter points predicted by SpaDiT and the real scatter points almost overlap, indicating that the genes predicted by SpaDiT are closer to the real genes.}
    \label{umap}
\end{figure*}
To fully demonstrate the superior ability of the SpaDiT method in gene expression prediction, especially its advantages in maintaining the global and local structural characteristics of gene expression data, we used UMAP technology for visualization analysis for conducting in-depth comparisons with other benchmark methods.

As shown in the \autoref{umap}, we conducted an analysis of ten different datasets. 
The results clearly show that the prediction results of the SpaDiT method (in orange) closely resemble the real gene expression data (in blue), with minimal perceptible deviation. 
This is in sharp contrast to the prediction results generated by several other methods, such as Tangram, scVI, SpaGE, stPlus, SpaOTsc, novoSpaRc, SpatialScope, and stDiff. 
Although the prediction results of these methods have their own focuses, compared with SpaDiT, they all fail to accurately capture the structural characteristics of real gene data and exhibit significant deviations. 
In addition, the UMAP analysis further underscores SpaDiT's superiority in maintaining data integrity, enabling it to accurately simulate complex biological information.

\subsection{\textbf{SpaDiT preserves the similarity between genes}}
%heatmap
\begin{figure*}
    \centering
    \includegraphics[width=\textwidth]{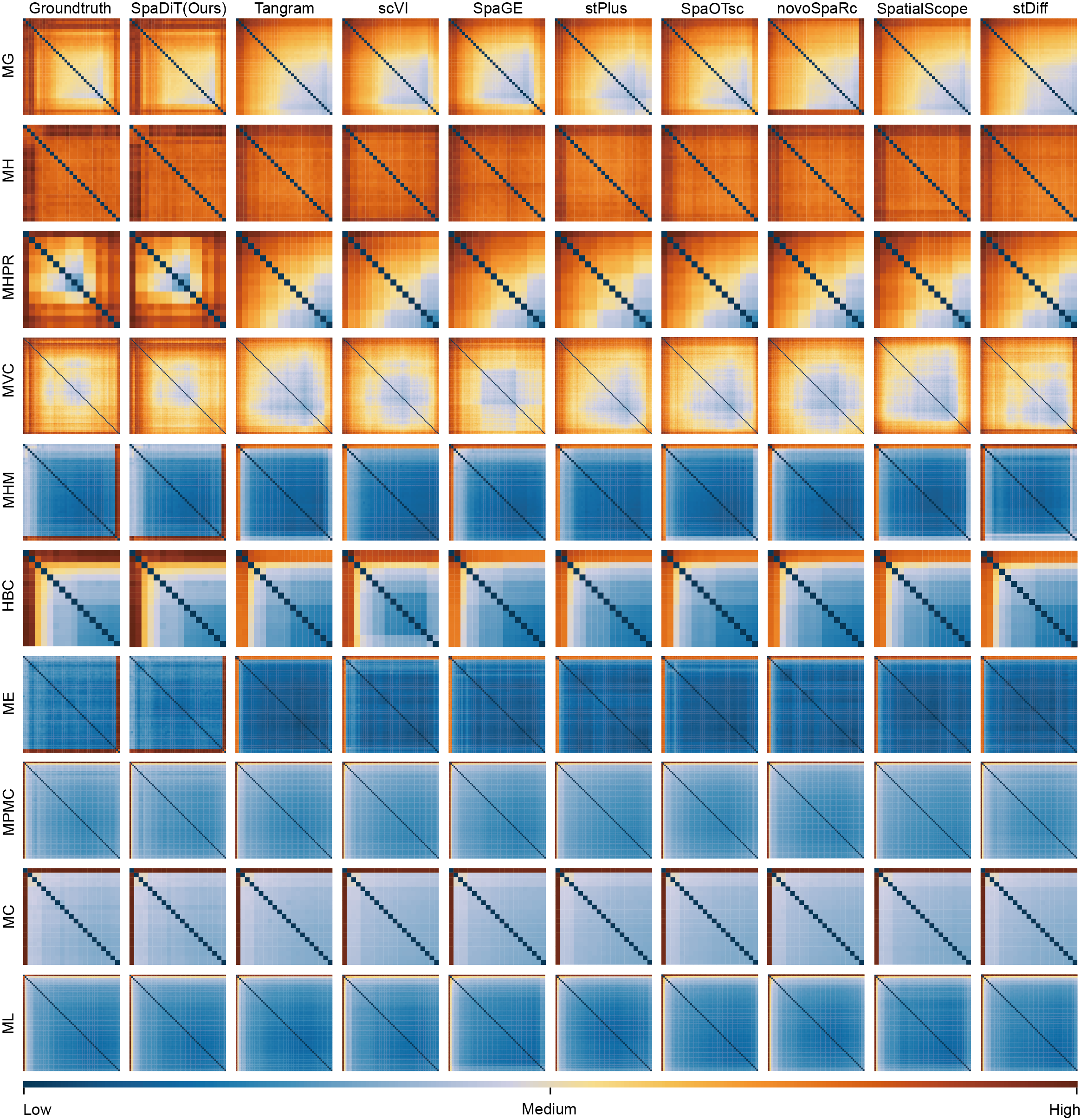}
    \caption{Visualization of the prediction performance of various baseline methods. The first column of the figure shows the results after clustering the true labels. The closer the predicted results of each method are to the true labels, the better the effect. The clustering effect of SpaDiT is closest to the true labels.}
    \label{heatmap}
\end{figure*}
To fully demonstrate the accuracy of the SpaDiT method in predicting gene expression, we employed hierarchical clustering to visualize the similarity between the predicted genes and the true gene labels, and compared the results with those from other benchmark methods.

First, we calculated the Euclidean distance between each pair of genes in the gene expression matrix predicted by each method to reflect the similarity of the expression patterns of two genes: the smaller the distance, the higher the similarity. After calculating the distance of all gene pairs, we used hierarchical clustering to sort these genes to ensure that the genes within the cluster show the greatest similarity. With this sorting, we can reorganize the rows and columns of the distance matrix so that similar genes are adjacent to each other in the heat map.

As shown in the \autoref{heatmap}, the first column of the figure visualizes the true gene labels after clustering. 
The closer the predicted gene heat map is to the true labels, the higher the prediction accuracy of the method. 
As evident from the figure, the prediction results of the SpaDiT method are very close to the true labels, demonstrating its high accuracy in predicting gene expression.

\subsection{\textbf{SpaDiT accurately predicts ST Spatial Patterns
}}
%Spatial
\begin{figure*}
    \centering
    \includegraphics[width=\linewidth]{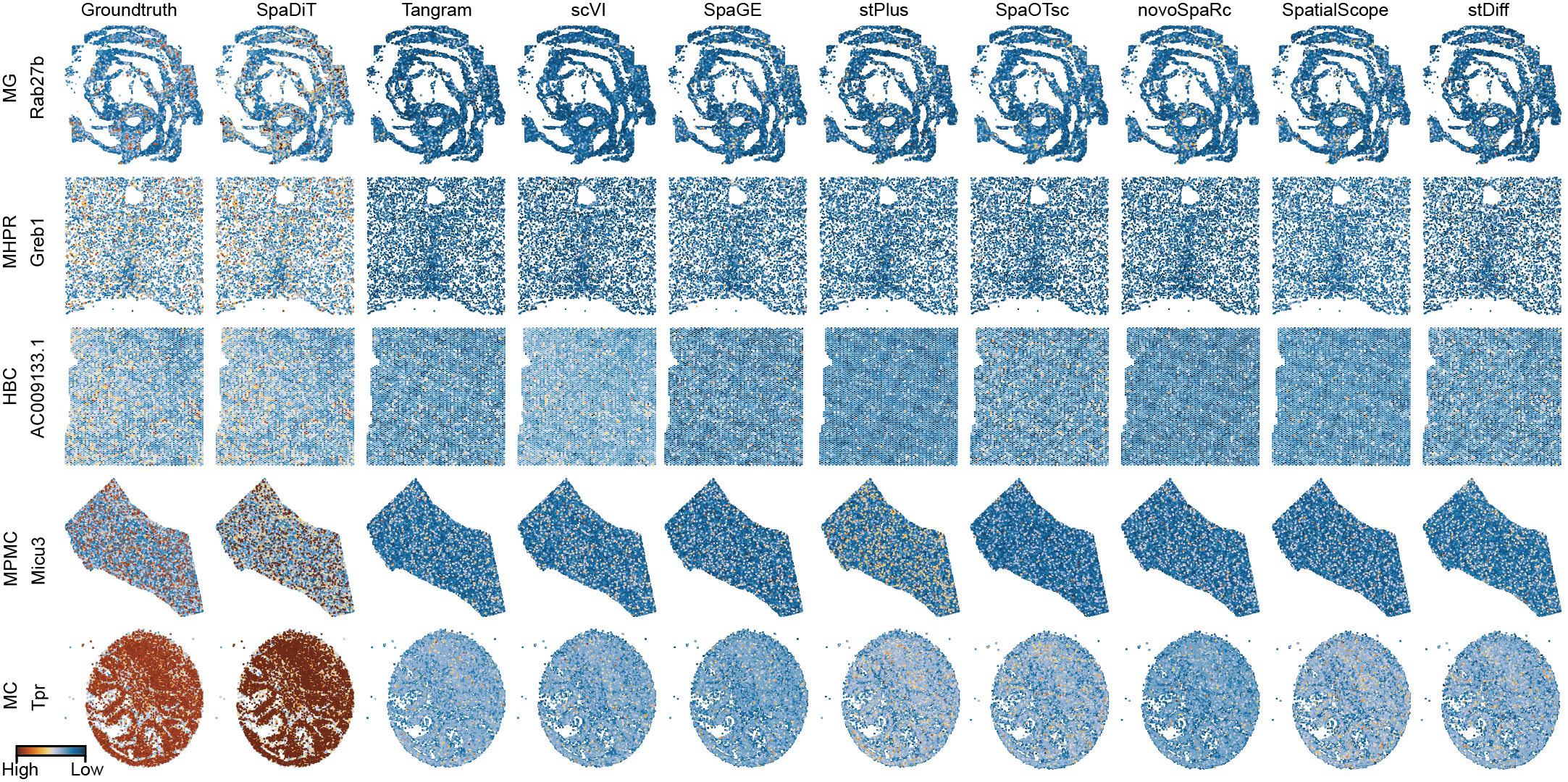}
    \caption{Predicted expression abundance of genes with known spatial patterns in four datasets. Each column corresponds to a gene with a clear spatial pattern. The first column represents the spatial pattern genes with true labels. Subsequent columns show the corresponding predicted expression patterns obtained by using SpaDiT, Tangram, scVI, SpaGE, stPlus, SpaOTsc, novoSpaRc, SpatialScope, and stDiff.}
    \label{spatial}
\end{figure*}
In addition to quantitatively evaluating the gene expression similarity between the true genes of ST and the genes predicted by ST, we also visually demonstrate the consistency of spatial patterns in \autoref{spatial}.

Due to limited space, we selected five datasets with clear spatial patterns: MG, MHPR, HBC, MPMC, and MC to illustrate the consistency of the spatial patterns between the genes predicted by the methods and the true labels. We display the predicted genes with the highest Pearson correlation coefficient (PCC) values in the datasets. The other five datasets not shown can be found in the \href{https://github.com/lllxxyyy-lxy/SpaDiT}{Supplementary Materials}.

As illustrated in \autoref{spatial}, in the MG dataset, SpaDiT restores the overall spatial pattern more accurately, followed by stDiff and stPlus, while the other methods show less obvious spatial contours in the upper right part. In the MHPR dataset, SpaDiT provides more accurate predictions in the middle part, while the high expression area and low expression area of other methods appear somewhat chaotic. In the HBC and MPMC datasets, all methods predict relatively accurate spatial patterns, but SpaDiT is the method with expression value predictions closest to the true labels. In the MC dataset, SpaDiT has a clear spatial recognition contour for the lower half, which is closest to the actual situation, while other methods are more blurred at the boundary.

\subsection{\textbf{Robustness evaluation of SpaDiT across various sampling rates
}}
\begin{figure*}
    \centering
    \includegraphics[width=\linewidth]{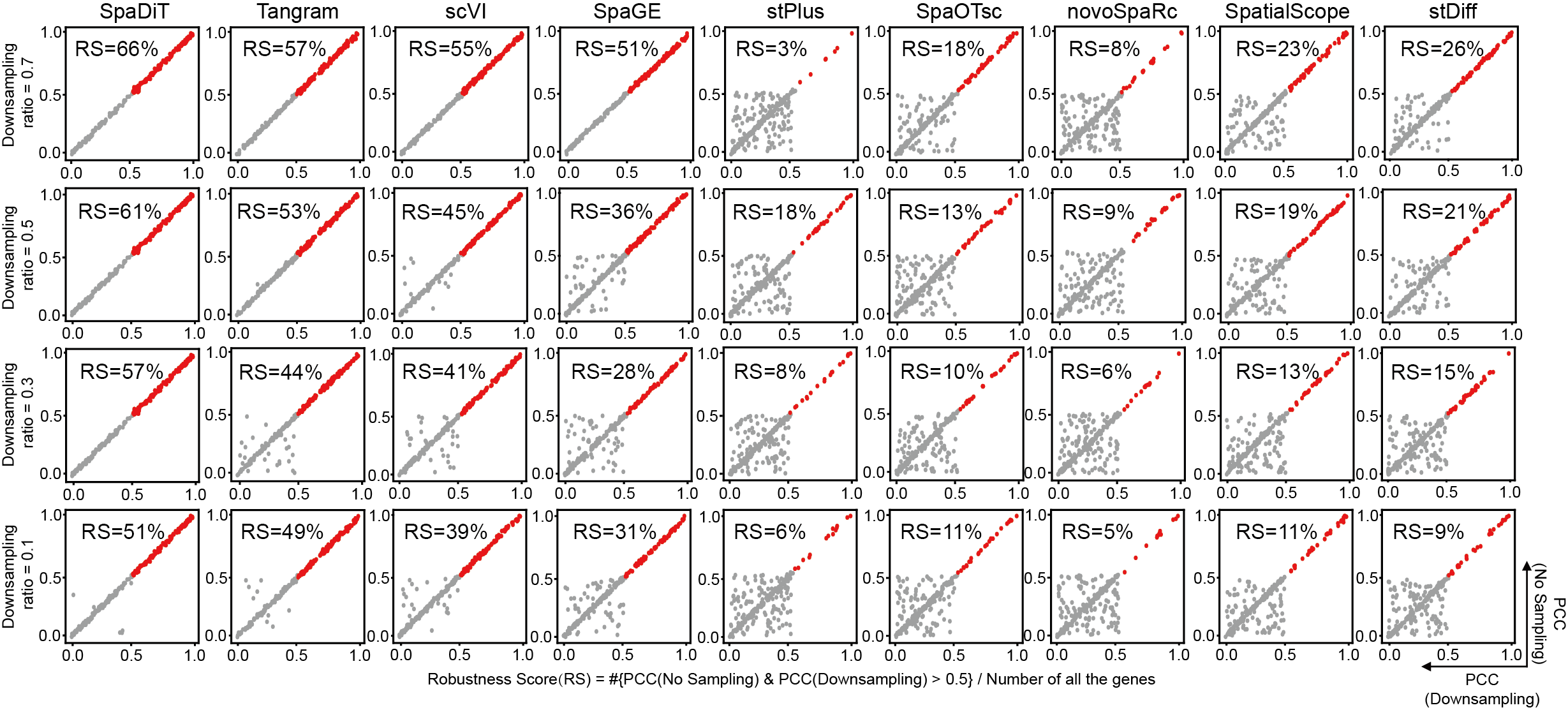}
    \caption{Robustness of prediction accuracy for original data and data with different downsampling rates for the MH dataset. PCC of the spatial distribution of transcripts predicted from the original data and the MH dataset at different downsampling ratios. The PCC values of red transcripts are greater than 0.5 for both the original data and the downsampled data. The proportion of red transcripts in all transcripts is defined as the “robustness score” (RS).}
    \label{rs}
\end{figure*}
In our study, the sparsity of the ten dataset pairs varies. 
Most of the datasets are highly sparse spatial transcriptomic data, except for the MH dataset, which has a sparsity of 6.7\%.
To test the ability of SpaDiT to resist data sparsity, we downsampled the expression matrix of MH's spatial transcriptomics data to simulate different high-sparse data. 
To quantify the stability of the SpaDiT and its ability to resist data sparsity, we counted the percentage of genes with a prediction accuracy (PCC) greater than 0.5 in both the original data and the downsampled data, defined as the Robustness Score (RS). 
As shown in the \autoref{rs}, the red points represent genes with a PCC value greater than 0.5, and the gray points represent genes with a PCC value less than 0.5. 
We tested different downsampling rates: 0.1, 0.3, 0.5, and 0.7, and found that the stability scores of all methods decreased with the increase of data sparsity, while the stability score of SpaDiT was always higher than that of other baseline methods.
In addition, we compared the changing trends of model performance under different sampling rates and different sparsity levels on ten datasets. For detailed results on other datasets, please refer to the \href{https://github.com/lllxxyyy-lxy/SpaDiT}{Supplementary Materials}.
\subsection{\textbf{Ablation studies: The impact of different modules of SpaDiT on model performance}}

%%%%%%%%%%%%%%%%%%%%%%%backbone
\begin{table*}[]
\centering
\caption{Result of different network backbone.}
\label{backbone}

\resizebox{\textwidth}{!}{%
\fontsize{3pt}{4pt}\selectfont
\setlength{\arrayrulewidth}{0.1mm}
\begin{tabular}{l|l|l|l|l|l}
\hline
 & \multicolumn{1}{c|}{MG}  & \multicolumn{1}{c|}{MH} & \multicolumn{1}{c|}{MHPR} & \multicolumn{1}{c|}{MVC} & \multicolumn{1}{c}{MHM} \\ \hline
Backbone w/Unet        & 0.454±0.011          & 0.453±0.011          & 0.477±0.013          & 0.482±0.011          & 0.466±0.011          \\
Backbone w/Mamba       & 0.477±0.008          & 0.471±0.026          & 0.475±0.014          & 0.474±0.011          & 0.461±0.102          \\
Backbone w/Transformer & \textbf{0.514±0.032} & \textbf{0.553±0.057} & \textbf{0.506±0.038} & \textbf{0.572±0.033} & \textbf{0.553±0.037} \\ \hline \noalign{\smallskip} \hline
 & \multicolumn{1}{c|}{HBC} & \multicolumn{1}{c|}{ME} & \multicolumn{1}{c|}{MPMC} & \multicolumn{1}{c|}{MC}  & \multicolumn{1}{c}{ML}  \\ \hline
Backbone w/Unet        & 0.478±0.012          & 0.470±0.010          & 0.458±0.013          & 0.470±0.010          & 0.487±0.013          \\
Backbone w/Mamba       & 0.462±0.086          & 0.489±0.051          & 0.421±0.022          & 0.478±0.015          & 0.488±0.021          \\
Backbone w/Transformer  & \textbf{0.613±0.024} & \textbf{0.589±0.060} & \textbf{0.488±0.033} & \textbf{0.564±0.026} & \textbf{0.619±0.024} \\ \hline
\end{tabular}%
}
\end{table*}

%%%%%%%%%%%%%%%%%%%%%%%%%%%%%%%%%%%%%%%%%Ablation3
\begin{table*}[]
\centering
\caption{Ablation study of Condition Embedding module and Latent Embedding module.}
\label{ablation3}
\resizebox{\textwidth}{!}{%
\fontsize{3pt}{4pt}\selectfont
\setlength{\arrayrulewidth}{0.1mm}
\begin{tabular}{l|c|c|c|c|c}
\hline
                         & MG          & MH          & MHPR        & MVC         & MHM         \\ \hline
\textbf{SpaDiT(Ours)} & \textbf{0.514±0.032} & \textbf{0.553±0.057} & \textbf{0.506±0.038} & \textbf{0.572±0.033} & \textbf{0.553±0.037} \\
w/o Flash-Attention            & 0.439±0.092 & 0.485±0.027 & 0.431±0.028 & 0.429±0.013 & 0.415±0.017 \\
w/o Condition $\psi$     & 0.383±0.094 & 0.336±0.115 & 0.404±0.161 & 0.394±0.066 & 0.318±0.013 \\
w/ Common Gene in $\psi$ & 0.483±0.126 & 0.503±0.008 & 0.437±0.125 & 0.533±0.161 & 0.489±0.088 \\
w/o Concat in $\phi$     & 0.462±0.093 & 0.501±0.140 & 0.432±0.020 & 0.489±0.076 & 0.485±0.042 \\ \hline\noalign{\smallskip} \hline
                         & HBC          & ME          & MPMC        & MC         & ML         \\  \hline 
\textbf{SpaDiT(Ours)}       & \textbf{0.613±0.024} & \textbf{0.589±0.060} & \textbf{0.488±0.033} & \textbf{0.564±0.026} & \textbf{0.619±0.024} \\
w/o Flash-Attention            & 0.431±0.034 & 0.422±0.021 & 0.425±0.013 & 0.438±0.019 & 0.459±0.033 \\
w/o Condition module: $\psi$     & 0.407±0.053 & 0.376±0.169 & 0.401±0.050 & 0.417±0.108 & 0.423±0.022 \\
w/ Common Gene in $\psi$ & 0.537±0.032 & 0.426±0.142 & 0.411±0.083 & 0.503±0.050 & 0.526±0.062 \\
w/o Concat in $\phi$     & 0.407±0.128 & 0.512±0.161 & 0.311±0.106 & 0.489±0.074 & 0.503±0.114 \\ \hline
\end{tabular}%
}
\end{table*}
As mentioned above, Condition Embedding and Backbone network in SpaDiT are the key parts of our proposed method. In order to verify the importance of these two parts, we conducted ablation experiments in this section.

For the backbone network part (\autoref{model}), we used three different network backbones and used AS as the evaluation indicator. As shown in the \autoref{backbone}, we compared three different network architectures: U-Net, Mamba, and Transformer. It is worth noting that the model using Transformer as the network backbone has the best performance, which further proves the importance of Transformer and verifies the superiority of SpaDiT.

In our proposed, SpaDiT, the main innovation involves using spatial transcriptomics (ST) and single-cell (SC) common genes to concatenate by gene in latent embedding. 
We use the known part (concatenated SC gene) to infer the gene expression of the unknown part (ST gene to be predicted). 
This approach enables the model to learn the similarity between different spots and cells across genes. 
Additionally, the Condition module utilizes the overall SC data as the prior condition to guide the model's generation process. 
To verify the effectiveness of the proposed method, we conducted ablation experiments on these two modules separately.

The specific experimental results are shown in \autoref{ablation3}.
First, for the Condition module ($\psi$), to verify the effectiveness of the Attention mechanism, we replaced Attention with a simple MLP (Part: w/o Flash-Attention). 
We found that the performance of the model dropped significantly across ten datasets. 
Further, to verify the effectiveness of the Condition module ($\psi$) (Part: w/o Condition $\psi$), we replaced the output of the entire part with a vector of all zeros. 
We observed that compared to replacing Attention, the performance of the model further declined. 
Additionally, to verify the effectiveness of the overall SC data as a priori conditions (Part: w/ Common Gene in $\psi$), we replaced the overall SC data with SC that only retained the common genes. 
We found that the performance also declined compared to the overall SC. 
Finally, to verify the effectiveness of the splicing of the common genes (Part: w/o Concat in $\phi$), we removed this part and found that the performance significantly declined. 
Therefore, we conclude that the method we proposed is highly effective.
In addition, we also tried using Condition modules with different Condition methods. For details, please refer to the \href{https://github.com/lllxxyyy-lxy/SpaDiT}{Supplementary Materials}.

\section{Discussion}
In this paper, we present SpaDiT, a novel approach to predict unmeasured genes in spatial transcriptomics (ST) data. 
Methodologically, SpaDiT is significantly different from existing ensemble techniques. 
While traditional approaches primarily enhance ST data by aligning ST data to similar cells within a reference scRNA-seq dataset, SpaDiT employs a diffusion-based generative model that utilizes the inherent relationships within the gene expression data. 
This approach enables it to precisely model and generate spatial gene expression patterns.

SpaDiT, as a conditional diffusion model, employs noise addition and denoising stages to learn complex relationships from scRNA-seq data. 
In the inference stage, SpaDiT incorporates raw ST data during the denoising process, resulting in accurate predictions of spatial gene expression. 
The application of diffusion models in genomics, especially transcriptomics, is relatively new, marking this as a largely unexplored area. 
We assessed SpaDiT using ten ST datasets, employing multiple metrics to evaluate performance, gene spatial structure, and gene similarity. 
The results show that SpaDiT not only maintains the intricate topology inherent in cell layout but also excels in accurately aligning predicted gene expression with actual data, demonstrating its robustness and accuracy in reproducing spatial patterns. 
These features highlight the utility of SpaDiT in enhancing the resolution and richness of ST data analysis.

Future research may combine SpaDiT's diffusion-based approach with traditional similarity-based methods to enhance the accuracy of ST data predictions. 
These advances may significantly improve the analysis and interpretation of ST data, potentially setting new standards in the field. 
It is important to acknowledge the potential limitations. 
For example, when ST data lack sufficient markers to accurately identify cell types, SpaDiT's efficacy may be diminished, similar to other methods. 
This is due to the reliance on existing gene expression signals to guide the prediction process, potentially resulting in inaccuracies if the initial data are too sparse or ambiguous. 
This underscores the need for improvements in handling datasets with limited information, ensuring that SpaDiT can adapt to various levels of data completeness and quality.

\begin{framed}  
\verb { {\textbf{Key Points} 
\begin{itemize}
    \item In this study, we propose SpaDiT, a deep learning method that utilizes a conditional diffusion generative model to synthesize scRNA-seq data and ST data to predict undetected genes.
    
    \item We utilize scRNA-seq as a prior condition and integrate it into the diffusion model through the attention mechanism to guide the model in learning the relationship between ST and scRNA-seq. At the same time, the common genes in ST and scRNA-seq are concatenated as the "token" input of the model, so that SpaDiT can learn multi-scale feature information and more accurately predict unknown genes.
    
    \item Our method was compared with competing methods on ten real ST and scRNA-seq datasets. The results show that, compared with the most advanced methods, our method demonstrates significant improvements in all five evaluation metrics in predicting gene expression. In addition, the genes predicted by our proposed SpaDiT effectively maintain high-dimensional similarity with the real labels, clearly restoring the spatial patterns between genes and the similarities between genes.
\end{itemize}
\end{framed} 

\section*{Acknowledgments}
The work was supported in part by the National Natural Science Foundation of China (62262069), in part by the Yunnan Fundamental Research Project (202301BF070001-019) and the Yunnan Talent Development Program - Youth Talent Project. 

\bibliographystyle{unsrt}
\bibliography{reference}

% \onecolumn
% \pagenumbering{arabic} 
% \setcounter{page}{1}

% \section*{\centering{\LARGE Appendix}} \vspace{15pt}

% \begin{bibunit}[unsrt] 
% \input{appendix.tex}
% \putbib[reference]
% \end{bibunit}

\end{document}